\documentclass{article}
\usepackage{spconf,amsmath,graphicx,hyperref}
\usepackage{amsmath}
\usepackage{amssymb}
\usepackage{xcolor}
\usepackage{graphicx}
\usepackage{booktabs}
\usepackage{array}
\usepackage{multirow}
\usepackage{colortbl}
\usepackage{caption}
\usepackage{hyperref}

\definecolor{mydarkblue}{RGB}{0, 51, 153}   % 深蓝色
\definecolor{wallcolor}{RGB}{174,199,232}
\definecolor{floorcolor}{RGB}{152,223,138}
\definecolor{cabinetcolor}{RGB}{31,119,180}
\definecolor{bedcolor}{RGB}{255,187,120}
\definecolor{chaircolor}{RGB}{188,189,34}
\definecolor{sofacolor}{RGB}{140,86,75}
\definecolor{tablecolor}{RGB}{255,152,150}
\definecolor{doorcolor}{RGB}{214,39,40}
\definecolor{windowcolor}{RGB}{197,176,213}
\definecolor{bookshelfcolor}{RGB}{148,103,189}
\definecolor{picturecolor}{RGB}{196,156,148}
\definecolor{countercolor}{RGB}{23,190,207}
\definecolor{deskcolor}{RGB}{247,182,210}
\definecolor{curtaincolor}{RGB}{219,219,141}
\definecolor{refrigeratorcolor}{RGB}{255,127,14}
\definecolor{showercolor}{RGB}{158,218,229}
\definecolor{toiletcolor}{RGB}{44,160,44}
\definecolor{sinkcolor}{RGB}{112,128,144}
\definecolor{bathtubcolor}{RGB}{227,119,194}
\definecolor{othercolor}{RGB}{82,84,163}
\definecolor{deltaRed}{RGB}{255,29,139}
\definecolor{deltaGreen}{RGB}{93,208,87}
\definecolor{resultgray}{RGB}{150,150,150}
\definecolor{s3disceilingcolor}{RGB}{0,255,0}
\definecolor{s3disfloorcolor}{RGB}{0,0,255}
\definecolor{s3diswallcolor}{RGB}{136,205,234}
\definecolor{s3disbeamcolor}{RGB}{255,255,0}
\definecolor{s3discolumncolor}{RGB}{218,116,218}
\definecolor{s3diswindowcolor}{RGB}{0,0,0}
\definecolor{s3disdoorcolor}{RGB}{162,147,126}
\definecolor{s3distablecolor}{RGB}{126,42,42}
\definecolor{s3dischaircolor}{RGB}{255,0,0}
\definecolor{s3dissofacolor}{RGB}{155,64,64}
\definecolor{s3disbookcasecolor}{RGB}{27,190,105}
\definecolor{s3disboardcolor}{RGB}{127,139,135}
\definecolor{s3discluttercolor}{RGB}{68,68,68}
\definecolor{ultralightgray}{RGB}{248, 248, 248}
\definecolor{tablegreen}{RGB}{225, 255, 225}   % 原来 207→235，更清淡
\definecolor{tableblue}{RGB}{230, 240, 255}   % 原来 169→210，232→240，更浅蓝
\definecolor{tablered}{RGB}{255, 235, 235}    % 原来 204→225，更浅粉红
\newcommand{\gain}[1]{\textcolor{deltaRed}{#1}}
\newcommand{\drop}[1]{\textcolor{deltaGreen}{#1}}

\newcommand{\best}[1]{\textcolor{black}{#1}}
\newcommand{\second}[1]{\textcolor{black}{\underline{#1}}}
\newcommand{\other}[1]{\textcolor{resultgray}{#1}}

\newcommand{\classlabel}[1]{\rotatebox{90}{#1}}
\newcommand{\classcolor}[1]{\textcolor{#1}{\rule{1.7ex}{1.7ex}}}

\hypersetup{
    colorlinks = true,                % 去掉方框，直接显示彩色文字
    urlcolor   = [rgb]{0, 0, 0.55},           % 深蓝（或 navy, MidnightBlue 等）
}

\title{Partition-invariant Tuning for 3D Scene Understanding}
\name{{\fontsize{10}{12.5}\selectfont
 Hongqiang Lin$^{1}$,Tianle Wang$^{2}$,Shuiwang Li$^{2}$,Dongxu Zhang$^{3}$ ,Yiding Sun$^{4}$,
Zihao Guo$^{5}$,Dongfu Yin$^{6}$\thanks{The author order was determined by a coin toss.}
}}

\address{{\fontsize{10.5}{11}\selectfont
$^{1}$ZJU,\hspace{1em}
$^{2}$GUT,\hspace{1em}
$^{3}$THU,\hspace{1em}
$^{4}$PKU,\hspace{1em}
$^{5}$XJTU,\hspace{1em}
$^{6}$GUANGMING LAB
}}
\begin{document}
\ninept
\maketitle
\begin{abstract}
  Scene-level point cloud understanding remains challenging due to diverse geometries and spatial layouts. While pre-trained 3D point cloud foundation models (PFMs) offer strong transferability, full fine-tuning (FFT) incurs substantial computational and storage costs. Parameter-efficient fine-tuning (PEFT) provides a promising alternative, but existing PEFT methods largely focus on object-level point clouds and overlook serialization-induced partition variations in large-scale scenes. To address this issue, we propose PointPiT, a partition-invariant tuning framework for scene-level point clouds. Specifically, a Scene-aware Structural Adapter (SSA) integrates local geometric patterns with global scene context to mitigate partition-induced representation shifts. Moreover, Gradient Subspace Optimization (GSO) selects informative and partition-stable update directions, suppressing partition-dependent variations during optimization. Extensive experiments across multiple scene-level benchmarks demonstrate that PointPiT achieves competitive or even superior performance to full fine-tuning with less than 1\% of backbone's parameters, while achieving consistent state-of-the-art performance among representative PEFT methods.
  % 只为这一个链接的文字设置为蓝色
\end{abstract}
\section{Introduction}
PFMs have demonstrated strong transferability across diverse 3D scene understanding tasks~\cite{wu2024pointtransformerv3,wu2025sonata,zhang2026pointcot,li2025scenesplat}, providing powerful pretrained representations for downstream adaptation~\cite{sun2026align,tang2026geometry,you2026gaussfusion,han2025rethinking}. Despite significant capabilities, downstream adaptation of large pretrained point cloud backbones still largely depends on FFT, incurring considerable computational and storage costs while requiring substantial task-specific supervision~\cite{zha2023idpt,liang2024pointgst,guo2026parameter}. To alleviate these limitations, PEFT has emerged as an efficient alternative, achieving competitive performance with only a small fraction of trainable parameters~\cite{zhang2026cmhanet,zhang2026diffusion,wang2026pointrft,han2025most}. However, existing methods largely focus on object-level point clouds, leaving large-scale scene adaptation underexplored.

When existing PEFT methods are extended to large-scale scene-level point clouds, the potential of pretrained representations is often not fully realized. Large-scale 3D scenes comprise dense point sets with substantial geometric heterogeneity, for which serialization-based grouping is essential to tractable computation~\cite{zhao2021pointtransformer,wu2022point,sun2026curve3d,wu2024pointtransformerv3}. Yet such grouping admits multiple valid partitions, exposing identical physical points to distinct local contexts. This variability can steer the limited adaptation capacity of PEFT toward partition-specific artifacts rather than task-relevant semantics.

Such partition sensitivity manifests differently across adaptation paradigms. As illustrated in Fig.~\ref{fig1}, FFT can largely reconcile partition-induced discrepancies through unrestricted backbone updates, whereas linear probing simply inherits the partition sensitivity of the frozen representation. More critically, LoRA~\cite{hu2022lora} adapts the backbone within a restricted update space yet lacks an explicit mechanism to distinguish task-relevant variation from partition-specific perturbations, thereby rendering its limited adaptation budget vulnerable to the latter. Such entanglement between task adaptation and partition-specific perturbations limits the effective exploitation of pretrained scene representations by existing PEFT methods~\cite{tang2024point,sun2026tri}.

In this paper, we propose PointPiT, a partition-invariant tuning framework
tailored for large-scale scene-level point clouds. PointPiT mitigates
partition sensitivity through both forward representation adaptation and backward
optimization. Specifically, the \emph{Scene-aware Structural Adapter (SSA)}
combines backbone-encoded geometric features with global scene context to mitigate
partition-induced representation shifts. \emph{Gradient Subspace Optimization (GSO)}
further characterizes gradient informativeness and partition-dependent
variation to identify a more stable update subspace, reducing sensitivity to
partition-dependent gradients during back-propagation.

Extensive experiments demonstrate the effectiveness and efficiency of
PointPiT. When integrated into Point Transformer V3
(PTv3)~\cite{wu2024pointtransformerv3}, PointPiT
reduces the number of trainable parameters by more than
99\% compared with full fine-tuning of the backbone,
while maintaining comparable or even better performance
across benchmarks. Our main contributions are summarized as follows:

\begin{figure}[!t]
  \centering
  \includegraphics[width=1\columnwidth]{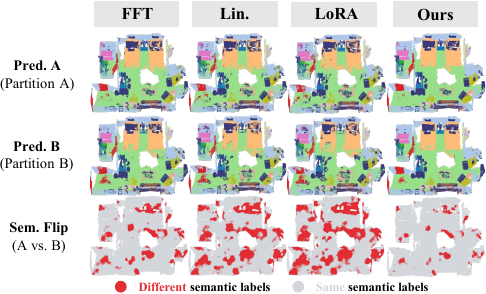} % Reduce the figure size so that it is slightly narrower than the column. Don't use precise values for figure width. This setup will avoid overfull boxes.
  \vspace{-12pt}
  \caption{Illustration of partition sensitivity on the ScanNet Val~\cite{dai2017scannet}.}
  \vspace{-12pt}
  \label{fig1}
\end{figure}

\begin{figure*}[!t]
  \centering
  \includegraphics[width=0.95\textwidth]{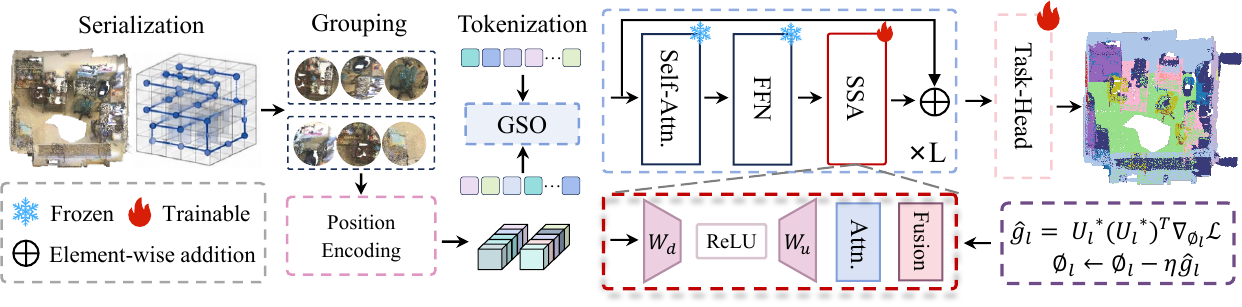} % Reduce the figure size so that it is slightly narrower than the column. Don't use precise values for figure width. This setup will avoid overfull boxes.
  \vspace{-12pt}
  \caption{\textbf{Overview of PointPiT.}
    Serialized scene points are grouped, position encoded, and embedded into tokens
    before entering the frozen backbone. PointPiT consists of a Scene-aware
    Structural Adapter (SSA) for forward representation adaptation and Gradient
    Subspace Optimization (GSO) for backward optimization. SSA is inserted into
    each backbone block to enhance scene-level structural representations, while GSO constrains gradient
    updates to partition-stable directions. During
    training, only SSA and the task head are updated.}
  \label{fig:pipeline}
\end{figure*}

\begin{itemize}
  \item We reveal partition sensitivity as a critical challenge for PEFT in large-scale 3D scenes, exposing the mismatch between scene serialization and parameter-efficient adaptation.

  \item We propose PointPiT, where SSA mitigates partition-induced representation
        shifts and GSO favors more stable gradient directions during optimization.

  \item Extensive experiments on multiple scene-level benchmarks demonstrate that PointPiT achieves SOTA performance among PEFT methods, while attaining performance comparable to FFT with substantially fewer trainable parameters.
\end{itemize}

\section{METHODOLOGY}

\subsection{Overview}

Given a large-scale point cloud scene
$\mathbf{X}\in\mathbb{R}^{N\times d}$,
where $N$ denotes the number of points and $d$ represents the input feature
dimension, PointPiT aims to efficiently adapt pretrained point cloud
foundation models to downstream scene understanding tasks. Building on the pioneer works~\cite{wu2024pointtransformerv3}, we define the point tokenization process as
$\mathcal{T}=\mathcal{E}\circ\mathcal{G}\circ\mathcal{S}$,
where $\mathcal{S}(\cdot)$ transforms unordered points into ordered
sequences, $\mathcal{G}(\cdot)$ partitions serialized points into local
groups, and $\mathcal{E}(\cdot)$ projects grouped points into feature tokens.
The initial token representation is therefore obtained as
$\mathbf{H}^{0}=\mathcal{T}(\mathbf{X})$.
The pretrained backbone consists of $L$ hierarchical blocks, each augmented
with SSA, whose forward process is defined as:
\begin{equation}
  \mathbf{H}^{l+1}
  =
  \mathcal{B}^{l}
  (\mathbf{H}^{l}),
  \quad l=0,1,\cdots,L-1,
\end{equation}
where $\mathcal{B}^{l}$ denotes the $l$-th backbone block.
To enable efficient adaptation for large-scale scenes, PointPiT introduces
two complementary components. The proposed \emph{Scene-aware Structural
  Adapter (SSA)} enhances forward representation adaptation by incorporating
both local geometric patterns and global scene context. Meanwhile,
\emph{Gradient Subspace Optimization (GSO)} optimizes the
backward update trajectory by preserving task-relevant adaptation directions.

\subsection{Scene-aware Structural Adapter}

Large-scale scenes exhibit diverse geometric structures, where serialized
local groups may suffer from limited contextual perception due to restricted
receptive fields. To improve scene-level adaptation, SSA combines geometric features with global scene context.

Given the intermediate representation
$\mathbf{H}^{l}\in\mathbb{R}^{N_l\times C_l}$,
SSA first adapts the geometry-aware features encoded by the
pretrained backbone through a lightweight bottleneck projection:
\begin{equation}
  \mathbf{H}_{loc}^{l}
  =
  \sigma\left(
  \mathbf{H}^{l}W_{d}^{l}
  \right)W_{u}^{l},
\end{equation}
where
$W_{d}^{l}\in\mathbb{R}^{C_l\times r}$ and
$W_{u}^{l}\in\mathbb{R}^{r\times C_l}$ denote the projection matrices,
$r$ represents the hidden dimension, and $\sigma(\cdot)$ denotes the
nonlinear activation function.

To incorporate global scene-level structural information, SSA
aggregates the input into a scene descriptor
$\mathbf{z}^{l}=\operatorname{AvgPool}(\mathbf{H}^{l})
  \in\mathbb{R}^{1\times C_l}$
and performs attention-weighted global aggregation:
\begin{equation}
  \mathbf{H}_{ctx}^{l}
  =
  \mathbf{1}_{N_l}
  \operatorname{Softmax}_{N_l}
  \left(
  \frac{\mathbf{H}^{l}(\mathbf{z}^{l}W_q^{l})^{T}}
  {\sqrt{C_l}}
  \right)^{T}
  (\mathbf{H}^{l}W_v^{l}),
\end{equation}
where $W_q^{l},W_v^{l}\in\mathbb{R}^{C_l\times C_l}$,
$\operatorname{Softmax}_{N_l}$ normalizes over points, and
$\mathbf{1}_{N_l}\in\mathbb{R}^{N_l\times1}$ broadcasts the
aggregated context to all points.

Finally, the local and contextual representations are adaptively fused
through a learnable gating mechanism:
\begin{equation}
  \widetilde{\mathbf{H}}^{l}
  =
  \mathbf{H}^{l}
  +
  \mathbf{g}^{l}\odot\mathbf{H}_{loc}^{l}
  +
  (1-\mathbf{g}^{l})\odot\mathbf{H}_{ctx}^{l},
\end{equation}
where
$\mathbf{g}^{l}=
  \operatorname{Sigmoid}(
  [\mathbf{H}_{loc}^{l},\mathbf{H}_{ctx}^{l}]W_g^{l}+b_g^{l})$
is a point-wise gate,
$W_g^{l}\in\mathbb{R}^{2C_l\times C_l}$,
and $\mathbf{g}^{l}\in(0,1)^{N_l\times C_l}$. $\odot$ represents element-wise multiplication.
$W_q^l$, $W_v^l$, and $W_g^l$ are implemented as rank-$t$ factorizations, with only their factor matrices trained.
Consequently, SSA establishes a structural adaptation pathway that mitigates
partition-induced representation shifts, while providing more consistent feature
representations for GSO to optimize gradient updates.

\begin{table*}[!t]
  \centering

  \caption{Semantic segmentation results on ScanNet Validation~\cite{dai2017scannet}. We report mean accuracy (mAcc), mean IoU (mIoU),
    per-class IoU (\%), and trainable parameters (M). The dec. and w/o dec. indicate Sonata with and without the decoder, respectively. Sonata (lin.) serves as the baseline for PEFT comparison. Among all PEFT methods, the best and second-best results are highlighted in black and underlined, respectively. The same notation applies to the following tables.}
  \vspace{-8pt}
  \label{tab:scannet}

  \fontsize{4.2}{5.3}\selectfont
  \setlength{\tabcolsep}{0.5pt}
  \renewcommand{\arraystretch}{1.02}

  \resizebox{\textwidth}{!}{%
    \begin{tabular}{l|ccc|*{20}{c}}
      \toprule

      % =========================================================
      % Row 1: ScanNet Val + class names
      % =========================================================

      \multirow[c]{2}{*}{%
        \raisebox{2.8ex}[0pt][0pt]{Method}%
      }                                &

      \multicolumn{3}{c|}{%
        \raisebox{4.0ex}[0pt][0pt]{ScanNet Val~\cite{dai2017scannet}}%
      }                                &

      \classlabel{wall}                &
      \classlabel{floor}               &
      \classlabel{cabinet}             &
      \classlabel{bed}                 &
      \classlabel{chair}               &
      \classlabel{sofa}                &
      \classlabel{table}               &
      \classlabel{door}                &
      \classlabel{window}              &
      \classlabel{books.}              &
      \classlabel{picture}             &
      \classlabel{counter}             &
      \classlabel{desk}                &
      \classlabel{curtain}             &
      \classlabel{refrig.}             &
      \classlabel{shower}              &
      \classlabel{toilet}              &
      \classlabel{sink}                &
      \classlabel{bathtub}             &
      \classlabel{other}
      \\[-2.8ex]

      \noalign{%
        \global\advance\aboverulesep by -0.5ex
        \global\advance\belowrulesep by  0.5ex
      }
      \cmidrule{2-4}
      \noalign{%
        \global\advance\aboverulesep by  0.5ex
        \global\advance\belowrulesep by -0.5ex
      }

      % =========================================================
      % Row 2: metrics + color blocks
      % =========================================================

                                       &
      \raisebox{1ex}[0pt][0pt]{Param.} &
      \raisebox{1ex}[0pt][0pt]{mAcc}   &
      \raisebox{1ex}[0pt][0pt]{mIoU}   &

      \classcolor{wallcolor}           &
      \classcolor{floorcolor}          &
      \classcolor{cabinetcolor}        &
      \classcolor{bedcolor}            &
      \classcolor{chaircolor}          &
      \classcolor{sofacolor}           &
      \classcolor{tablecolor}          &
      \classcolor{doorcolor}           &
      \classcolor{windowcolor}         &
      \classcolor{bookshelfcolor}      &
      \classcolor{picturecolor}        &
      \classcolor{countercolor}        &
      \classcolor{deskcolor}           &
      \classcolor{curtaincolor}        &
      \classcolor{refrigeratorcolor}   &
      \classcolor{showercolor}         &
      \classcolor{toiletcolor}         &
      \classcolor{sinkcolor}           &
      \classcolor{bathtubcolor}        &
      \classcolor{othercolor}
      \\

      % =========================================================
      % Full Fine-tuning
      % =========================================================

      \specialrule{\lightrulewidth}{0.4ex}{0.45ex}

      \rowcolor{gray!10}
      \multicolumn{24}{c}{%
        \fontsize{4.3}{4.7}\selectfont\itshape
        Full\hspace{3pt}Fine-tuning
      }                                                                                                                              \\

      \specialrule{\lightrulewidth}{0.45ex}{0.4ex}

      PTv3-PPT (sup.)~\cite{wu2024towards}
                                       & 124.8
                                       & \other{85.9}
                                       & \other{78.6}
                                       & \other{87.6}                & \other{96.3}  & \other{75.4}  & \other{85.3}  & \other{88.9}
                                       & \other{83.9}                & \other{77.3}  & \other{69.0}  & \other{70.0}  & \other{75.3}
                                       & \other{66.0}                & \other{72.6}  & \other{69.9}  & \other{79.0}  & \other{77.7}
                                       & \other{79.4}                & \other{93.7}  & \other{75.7}  & \other{85.8}  & \other{63.2}
      \\

      Sonata (dec.)~\cite{wu2025sonata}
                                       & 124.8
                                       & \other{86.1}
                                       & \other{79.4}
                                       & \other{88.3}                & \other{96.3}  & \other{76.3}  & \other{86.3}  & \other{89.0}
                                       & \other{84.8}                & \other{78.4}  & \other{70.2}  & \other{72.2}  & \other{76.2}
                                       & \other{67.4}                & \other{74.1}  & \other{70.4}  & \other{80.1}  & \other{78.8}
                                       & \other{79.8}                & \other{93.6}  & \other{76.6}  & \other{85.4}  & \other{63.8}
      \\

      Sonata (w/o dec.)~\cite{wu2025sonata}
                                       & 108.5 (100\%)
                                       & \other{86.1}
                                       & \other{78.9}
                                       & \other{87.6}                & \other{96.2}  & \other{76.1}  & \other{85.3}  & \other{88.8}
                                       & \other{84.6}                & \other{77.8}  & \other{69.8}  & \other{70.8}  & \other{75.5}
                                       & \other{66.6}                & \other{73.7}  & \other{69.9}  & \other{79.4}  & \other{77.6}
                                       & \other{80.0}                & \other{93.0}  & \other{76.1}  & \other{85.8}  & \other{63.4}
      \\

      % =========================================================
      % General PEFT Methods
      % =========================================================

      \specialrule{\lightrulewidth}{0.4ex}{0.45ex}

      \rowcolor{gray!10}
      \multicolumn{24}{c}{%
        \fontsize{4.3}{4.7}\selectfont\itshape
        General\hspace{2pt}PEFT\hspace{2pt}Methods
      }                                                                                                                              \\

      \specialrule{\lightrulewidth}{0.45ex}{0.4ex}

      Sonata (lin.)
                                       & 0.02 (0.02\%)
                                       & \other{83.1}
                                       & \other{72.2}
                                       & \other{82.8}                & \other{93.5}  & \other{70.4}  & \other{82.2}  & \other{85.8}
                                       & \other{78.6}                & \other{71.5}  & \other{58.4}  & \other{63.0}  & \other{69.3}
                                       & \other{55.0}                & \other{65.4}  & \other{59.8}  & \other{71.5}  & \other{71.0}
                                       & \other{72.7}                & \other{91.3}  & \other{68.3}  & \other{80.9}  & \other{52.6}
      \\

      + Adapter~\cite{houlsby2019parameter}
                                       & 1.90 (1.76\%)
                                       & \other{85.1} (\gain{+2.0})
                                       & \other{76.8} (\gain{+4.6})
                                       & \other{86.0}                & \best{97.0}   & \other{74.5}  & \other{84.7}  & \other{87.4}
                                       & \other{82.4}                & \other{77.3}  & \second{68.5} & \other{68.3}  & \other{72.0}
                                       & \other{62.0}                & \other{69.9}  & \other{67.6}  & \other{76.8}  & \other{75.6}
                                       & \other{78.2}                & \other{91.8}  & \other{72.6}  & \other{84.1}  & \other{59.3}
      \\

      + Prefix Tuning~\cite{li2021prefix}
                                       & 0.08 (0.08\%)
                                       & \other{84.2} (\gain{+1.1})
                                       & \other{73.5} (\gain{+1.3})
                                       & \other{84.7}                & \other{96.3}  & \other{70.0}  & \other{81.9}  & \other{85.9}
                                       & \other{80.1}                & \other{72.4}  & \other{61.8}  & \other{64.0}  & \other{69.8}
                                       & \other{57.5}                & \other{66.5}  & \other{63.7}  & \other{72.8}  & \other{72.6}
                                       & \other{74.7}                & \other{92.7}  & \other{69.3}  & \other{80.7}  & \other{52.6}
      \\

      + LoRA~\cite{hu2022lora}
                                       & 0.89 (0.82\%)
                                       & \other{85.2} (\gain{+2.1})
                                       & \other{76.3} (\gain{+4.1})
                                       & \other{85.5}                & \second{96.8} & \other{72.6}  & \other{84.5}  & \other{85.9}
                                       & \other{81.9}                & \other{74.3}  & \other{65.5}  & \other{67.8}  & \other{71.1}
                                       & \other{61.9}                & \other{71.2}  & \other{67.1}  & \other{77.8}  & \other{75.9}
                                       & \other{76.7}                & \other{92.7}  & \other{72.8}  & \other{84.3}  & \other{59.7}
      \\

      % =========================================================
      % PEFT Methods for Point Cloud
      % =========================================================

      \specialrule{\lightrulewidth}{0.4ex}{0.45ex}

      \rowcolor{gray!10}
      \multicolumn{24}{c}{%
        \fontsize{4.3}{4.7}\selectfont\itshape
        PEFT\hspace{2pt}Methods\hspace{2pt}for\hspace{2pt}Point\hspace{2pt}Cloud
      }                                                                                                                              \\

      \specialrule{\lightrulewidth}{0.45ex}{0.4ex}

      + IDPT~\cite{zha2023idpt}
                                       & 2.61 (2.42\%)
                                       & \other{82.9} (\drop{-0.2})
                                       & \other{72.6} (\gain{+0.4})
                                       & \other{84.4}                & \other{95.2}  & \other{69.1}  & \other{81.7}  & \other{85.2}
                                       & \other{79.4}                & \other{71.1}  & \other{60.5}  & \other{62.0}  & \other{68.0}
                                       & \other{57.9}                & \other{65.7}  & \other{61.7}  & \other{72.8}  & \other{70.8}
                                       & \other{73.1}                & \other{91.0}  & \other{69.4}  & \other{81.0}  & \other{52.0}
      \\

      + DAPT~\cite{zhou2024dapt}
                                       & 1.14 (1.06\%)
                                       & \second{86.5} (\gain{+3.4})
                                       & \other{77.7} (\gain{+5.5})
                                       & \other{86.7}                & \other{95.9}  & \other{75.2}  & \other{85.2}  & \other{87.8}
                                       & \second{83.6}               & \other{76.8}  & \other{68.2}  & \other{69.2}  & \other{73.8}
                                       & \other{64.1}                & \other{71.1}  & \other{68.0}  & \other{78.3}  & \second{76.9}
                                       & \second{79.6}               & \best{93.9}   & \other{74.2}  & \second{84.9} & \other{60.6}
      \\

      + PointGST~\cite{liang2024pointgst}
                                       & 1.05 (0.97\%)
                                       & \other{85.8} (\gain{+2.7})
                                       & \other{77.7} (\gain{+5.5})
                                       & \other{85.7}                & \other{96.4}  & \other{73.9}  & \other{84.9}  & \best{89.3}
                                       & \other{82.5}                & \other{76.9}  & \other{68.3}  & \other{69.0}  & \other{73.8}
                                       & \second{65.4}               & \other{71.5}  & \other{67.9}  & \other{78.2}  & \second{76.9}
                                       & \other{79.5}                & \second{93.8} & \other{74.4}  & \other{84.3}  & \other{61.4}
      \\

      + PointTPA~\cite{Liu_2026_CVPR}
                                       & 1.18 (1.09\%)
                                       & \other{86.3} (\gain{+3.2})
                                       & \second{78.4} (\gain{+6.2})
                                       & \best{88.0}                 & \other{96.6}  & \second{75.3} & \second{85.9} & \other{88.5}
                                       & \other{83.5}                & \best{78.1}   & \other{68.2}  & \second{70.3} & \best{75.1}
                                       & \other{65.1}                & \second{72.6} & \second{69.0} & \second{78.9} & \other{76.3}
                                       & \second{79.6}               & \other{93.7}  & \second{75.2} & \other{84.5}  & \best{63.6}
      \\

      \rowcolor{gray!10}
      + PointPiT (ours)
                                       & 1.08 (0.99\%)
                                       & \best{86.7} (\gain{+3.6})
                                       & \best{78.9} (\gain{+6.7})
                                       & \second{87.3}               & \other{96.5}  & \best{75.9}   & \best{86.3}   & \second{89.1}
                                       & \best{84.0}                 & \second{77.9} & \best{69.6}   & \best{71.2}   & \second{75.0}
                                       & \best{66.3}                 & \best{73.2}   & \best{70.1}   & \best{79.2}   & \best{78.0}
                                       & \best{79.8}                 & \second{93.8} & \best{75.9}   & \best{85.6}   & \second{63.3}
      \\

      \noalign{\vskip -0.5ex}
      \bottomrule
    \end{tabular}%
  }

  \vspace{-6pt}
\end{table*}

\begin{table*}[!t]
  \centering

  \caption{
    Semantic segmentation results on S3DIS Area 5~\cite{Armeni_2016_CVPR},
    S3DIS 6-fold~\cite{Armeni_2016_CVPR},
    ScanNet200 Validation~\cite{rozenberszki2022language},
    and ScanNet++ Validation~\cite{yeshwanthliu2023scannetpp} sets.
    We report mean accuracy (mAcc), overall accuracy (allAcc),
    and mean IoU (mIoU). Trainable parameters are consistent with Table~\ref{tab:scannet}.
  }
  \vspace{-8pt}
  \label{tab:indoor_seg}

  \fontsize{5.0}{6.0}\selectfont
  \setlength{\tabcolsep}{1pt}
  \renewcommand{\arraystretch}{1.05}

  \resizebox{\textwidth}{!}{%
    \begin{tabular}{l|ccc|ccc|ccc|ccc}
      \toprule

      % =========================================================
      % Header
      % =========================================================

      \multirow[c]{2}{*}{Method}
       &
      \multicolumn{3}{c|}{%
        \raisebox{0.5ex}[0pt][0pt]{S3DIS Area 5~\cite{Armeni_2016_CVPR}}
      }
       &
      \multicolumn{3}{c|}{%
        \raisebox{0.5ex}[0pt][0pt]{S3DIS 6-fold~\cite{Armeni_2016_CVPR}}
      }
       &
      \multicolumn{3}{c|}{%
        \raisebox{0.5ex}[0pt][0pt]{ScanNet200 Val~\cite{rozenberszki2022language}}
      }
       &
      \multicolumn{3}{c}{%
        \raisebox{0.5ex}[0pt][0pt]{ScanNet++ Val~\cite{yeshwanthliu2023scannetpp}}
      }
      \\

      \cline{2-4}
      \cline{5-7}
      \cline{8-10}
      \cline{11-13}

       &
      \raisebox{-0.45ex}{mAcc}
       &
      \raisebox{-0.45ex}{allAcc}
       &
      \raisebox{-0.45ex}{mIoU}
       &
      \raisebox{-0.45ex}{mAcc}
       &
      \raisebox{-0.45ex}{allAcc}
       &
      \raisebox{-0.45ex}{mIoU}
       &
      \raisebox{-0.45ex}{mAcc}
       &
      \raisebox{-0.45ex}{allAcc}
       &
      \raisebox{-0.45ex}{mIoU}
       &
      \raisebox{-0.45ex}{mAcc}
       &
      \raisebox{-0.45ex}{allAcc}
       &
      \raisebox{-0.45ex}{mIoU}
      \\

      \specialrule{\lightrulewidth}{0.7ex}{0.45ex}

      % =========================================================
      % Full Fine-tuning
      % =========================================================

      \rowcolor{gray!10}
      \multicolumn{13}{c}{
        \fontsize{5.0}{5.5}\selectfont\itshape
        Full\hspace{3pt}Fine-tuning
      }
      \\
      \specialrule{\lightrulewidth}{0.4ex}{0.45ex}

      PTv3-PPT (sup.)~\cite{wu2024towards}
       &
      \other{80.1}
       &
      \other{92.0}
       &
      \other{74.3}
       &
      \other{88.6}
       &
      \other{92.7}
       &
      \other{80.5}
       &
      \other{45.9}
       &
      \other{83.8}
       &
      \other{35.9}
       &
      \other{55.7}
       &
      \other{86.4}
       &
      \other{43.3}
      \\

      Sonata (dec.)~\cite{wu2025sonata}
       &
      \other{81.6}
       &
      \other{93.0}
       &
      \other{76.0}
       &
      \other{89.9}
       &
      \other{93.3}
       &
      \other{82.3}
       &
      \other{46.5}
       &
      \other{84.4}
       &
      \other{36.8}
       &
      \other{55.8}
       &
      \other{86.6}
       &
      \other{43.7}
      \\

      Sonata (w/o dec.)~\cite{wu2025sonata}
       &
      \other{80.3}
       &
      \other{93.1}
       &
      \other{74.5}
       &
      \other{87.3}
       &
      \other{92.3}
       &
      \other{79.5}
       &
      \other{47.8}
       &
      \other{83.7}
       &
      \other{37.3}
       &
      \other{50.7}
       &
      \other{86.3}
       &
      \other{41.8}
      \\

      % =========================================================
      % General PEFT
      % =========================================================

      \specialrule{\lightrulewidth}{0.4ex}{0.45ex}

      \rowcolor{gray!10}
      \multicolumn{13}{c}{
        \fontsize{5.0}{5.5}\selectfont\itshape
        General\hspace{2pt}PEFT\hspace{2pt}Methods
      }
      \\

      \specialrule{\lightrulewidth}{0.4ex}{0.45ex}

      Sonata (lin.)
       &
      \other{80.9}
       &
      \other{90.8}
       &
      \other{73.0}
       &
      \other{87.4}
       &
      \other{90.8}
       &
      \other{76.5}
       &
      \other{41.6}
       &
      \other{81.2}
       &
      \other{29.3}
       &
      \other{49.4}
       &
      \other{84.9}
       &
      \other{36.5}
      \\

      + Adapter~\cite{houlsby2019parameter}
       &
      \other{81.7}(\gain{+0.8})
       &
      \other{91.6}(\gain{+0.8})
       &
      \other{73.8}(\gain{+0.8})
       &
      \other{87.5}(\gain{+0.1})
       &
      \other{91.4}(\gain{+0.6})
       &
      \other{76.4}(\drop{-0.1})
       &
      \other{45.7}(\gain{+4.1})
       &
      \other{82.5}(\gain{+1.3})
       &
      \other{33.6}(\gain{+4.3})
       &
      \second{52.9}(\gain{+3.5})
       &
      \other{86.1}(\gain{+1.2})
       &
      \other{39.9}(\gain{+3.4})
      \\

      + Prefix Tuning~\cite{li2021prefix}
       &
      \best{82.5}(\gain{+1.6})
       &
      \other{91.0}(\gain{+0.2})
       &
      \other{73.4}(\gain{+0.4})
       &
      \other{86.5}(\drop{-0.9})
       &
      \other{90.5}(\drop{-0.3})
       &
      \other{73.7}(\drop{-2.8})
       &
      \other{44.4}(\gain{+2.8})
       &
      \other{81.6}(\gain{+0.4})
       &
      \other{31.4}(\gain{+2.1})
       &
      \other{49.6}(\gain{+0.2})
       &
      \other{84.2}(\drop{-0.7})
       &
      \other{36.8}(\gain{+0.3})
      \\

      + LoRA~\cite{hu2022lora}
       &
      \other{81.1}(\gain{+0.2})
       &
      \best{92.5}(\gain{+1.7})
       &
      \second{74.0}(\gain{+1.0})
       &
      \other{87.8}(\gain{+0.4})
       &
      \other{91.2}(\gain{+0.4})
       &
      \other{77.4}(\gain{+0.9})
       &
      \other{45.5}(\gain{+3.9})
       &
      \other{82.7}(\gain{+1.5})
       &
      \other{33.6}(\gain{+4.3})
       &
      \other{51.4}(\gain{+2.0})
       &
      \other{85.3}(\gain{+0.4})
       &
      \other{38.7}(\gain{+2.2})
      \\

      % =========================================================
      % Point Cloud PEFT
      % =========================================================

      \specialrule{\lightrulewidth}{0.4ex}{0.45ex}

      \rowcolor{gray!10}
      \multicolumn{13}{c}{
        \fontsize{5.0}{5.5}\selectfont\itshape
        PEFT\hspace{2pt}Methods\hspace{2pt}for\hspace{2pt}Point\hspace{2pt}Cloud
      }
      \\

      \specialrule{\lightrulewidth}{0.4ex}{0.45ex}

      + IDPT~\cite{zha2023idpt}
       &
      \other{81.2}(\gain{+0.3})
       &
      \other{90.5}(\drop{-0.3})
       &
      \other{72.0}(\drop{-1.0})
       &
      \other{86.8}(\drop{-0.6})
       &
      \other{90.6}(\drop{-0.2})
       &
      \other{75.2}(\drop{-1.3})
       &
      \other{41.9}(\gain{+0.3})
       &
      \other{81.0}(\drop{-0.2})
       &
      \other{29.0}(\drop{-0.3})
       &
      \other{50.0}(\gain{+0.6})
       &
      \other{84.1}(\drop{-0.8})
       &
      \other{36.6}(\gain{+0.1})
      \\

      + DAPT~\cite{zhou2024dapt}
       &
      \other{81.2}(\gain{+0.3})
       &
      \other{92.7}(\gain{+1.9})
       &
      \other{74.6}(\gain{+1.6})
       &
      \other{88.1}(\gain{+0.7})
       &
      \other{91.8}(\gain{+1.0})
       &
      \other{78.2}(\gain{+1.7})
       &
      \other{45.8}(\gain{+4.2})
       &
      \other{82.9}(\gain{+1.7})
       &
      \other{34.2}(\gain{+4.9})
       &
      \other{50.9}(\gain{+1.5})
       &
      \other{85.2}(\gain{+0.3})
       &
      \other{39.6}(\gain{+3.1})
      \\

      + PointGST~\cite{liang2024pointgst}
       &
      \other{81.4}(\gain{+0.5})
       &
      \other{92.9}(\gain{+2.1})
       &
      \second{74.0}(\gain{+1.0})
       &
      \other{88.6}(\gain{+1.2})
       &
      \other{92.2}(\gain{+1.4})
       &
      \other{78.9}(\gain{+2.4})
       &
      \other{46.5}(\gain{+4.9})
       &
      \other{83.2}(\gain{+2.0})
       &
      \other{35.0}(\gain{+5.7})
       &
      \other{52.0}(\gain{+2.6})
       &
      \other{85.9}(\gain{+1.0})
       &
      \other{40.0}(\gain{+3.5})
      \\

      + PointTPA~\cite{Liu_2026_CVPR}
       &
      \other{81.7}(\gain{+0.8})
       &
      \other{92.9}(\gain{+2.1})
       &
      \other{74.9}(\gain{+1.9})
       &
      \second{88.9}(\gain{+1.5})
       &
      \second{92.5}(\gain{+1.7})
       &
      \second{79.4}(\gain{+2.9})
       &
      \second{47.2}(\gain{+5.6})
       &
      \second{83.8}(\gain{+2.6})
       &
      \second{35.8}(\gain{+6.5})
       &
      \second{52.9}(\gain{+3.5})
       &
      \second{86.1}(\gain{+1.2})
       &
      \second{40.9}(\gain{+4.4})
      \\

      \rowcolor{gray!10}
      + PointPiT (ours)
       &
      \second{82.3}(\gain{+1.4})
       &
      \best{93.3}(\gain{+2.5})
       &
      \best{75.0}(\gain{+2.0})
       &
      \best{90.2}(\gain{+2.8})
       &
      \best{93.6}(\gain{+2.8})
       &
      \best{80.6}(\gain{+4.1})
       &
      \best{49.0}(\gain{+7.4})
       &
      \best{84.8}(\gain{+3.6})
       &
      \best{38.1}(\gain{+8.8})
       &
      \best{56.4}(\gain{+7.0})
       &
      \best{87.0}(\gain{+2.1})
       &
      \best{44.2}(\gain{+7.7})
      \\

      \noalign{\vskip -0.5ex}
      \bottomrule
    \end{tabular}
  }

  \vspace{-8pt}

\end{table*}

\subsection{Gradient Subspace Optimization}

Although SSA mitigates partition-induced variation in forward representations,
backward optimization remains susceptible to partition-dependent gradient
variations. GSO therefore favors informative and partition-stable update directions.
For each trainable module with parameters
$\phi_l\in\mathbb{R}^{D_l}$, we maintain a rolling buffer of $B$
scenes, each with $K$ valid partitions. Let
$\mathbf{g}_{bk}$ denote the stored gradient for the $k$-th
partition of scene $b$, and
$\bar{\mathbf{g}}_b=K^{-1}\sum_{k=1}^{K}\mathbf{g}_{bk}$.
The gradient second moment and partition covariance are then
estimated from these stored gradients as follows:
\begin{equation}
  \begin{aligned}
    \mathbf{F}_l
     & =\frac{1}{BK}\sum_{b,k}
    \mathbf{g}_{bk}\mathbf{g}_{bk}^{T}, \\
    \mathbf{G}_{p,l}
     & =\frac{1}{BK}\sum_{b,k}
    (\mathbf{g}_{bk}-\bar{\mathbf{g}}_b)
    (\mathbf{g}_{bk}-\bar{\mathbf{g}}_b)^{T}.
  \end{aligned}
\end{equation}
GSO seeks a subspace that balances informative update directions against
partition-dependent gradient variations:
\begin{equation}
  \mathbf{U}_l^{*}
  =
  \operatorname*{argmax}_{\substack{
        0\le s\le q,\;
        \mathbf{U}\in\mathbb{R}^{D_l\times s}\\
        \mathbf{U}^{T}\mathbf{U}=\mathbf{I}_{s}
      }}
  \operatorname{Tr}
  \left[
    \mathbf{U}^{T}
    (\mathbf{F}_l-\lambda\mathbf{G}_{p,l})
    \mathbf{U}
    \right],
\end{equation}
where $\lambda$ controls partition regularization and the subspace
dimension is capped at $q$.

To avoid constructing $D_l\times D_l$ matrices, we stack the stored
gradients into $\mathbf{Z}_l\in\mathbb{R}^{D_l\times BK}$ and define
$\mathbf{C}=\mathbf{I}_{B}\otimes
  (\mathbf{I}_{K}-\mathbf{1}_{K}\mathbf{1}_{K}^{T}/K)$.
With the thin SVD $\mathbf{Z}_l=\mathbf{Q}_l\mathbf{\Sigma}_l
  \mathbf{V}_l^{T}$, the subspace is recovered from the leading
eigenvectors of
\begin{equation}
  \mathbf{M}_l=
  \frac{1}{BK}\mathbf{\Sigma}_l\mathbf{V}_l^{T}
  (\mathbf{I}-\lambda\mathbf{C})
  \mathbf{V}_l\mathbf{\Sigma}_l.
\end{equation}
Specifically, $\mathbf{U}_l^{*}=\mathbf{Q}_l\mathbf{E}_l$,
where $\mathbf{E}_l$ contains at most $q$ eigenvectors with positive
eigenvalues. Thus, $q_{\mathrm{eff}}=
  \min(q,n_{+}(\mathbf{M}_l))$,
where $n_{+}(\mathbf{M}_l)$ denotes the number of positive
eigenvalues. The eigenproblem has dimension at most
$BK$, rather than $D_l$.

During back-propagation, GSO projects each module's gradient onto
the subspace spanned by $\mathbf U_l^*$, favoring directions
with lower partition-dependent variation:
\begin{equation}
  \hat{\mathbf g}_l
  =
  \mathbf U_l^*(\mathbf U_l^*)^T
  \nabla_{\phi_l}\mathcal L.
\end{equation}
The parameters are then updated by
$\phi_l\leftarrow\phi_l-\eta\hat{\mathbf g}_l$,
where $\eta$ is the learning rate.
This restricts updates to the selected subspace without changing
the set of trainable parameters. Consequently, GSO reduces
sensitivity to partition-dependent gradient variations and complements
the representation adaptation of SSA.
\begin{figure*}[!t]
  \centering
  \includegraphics[width=1\textwidth]{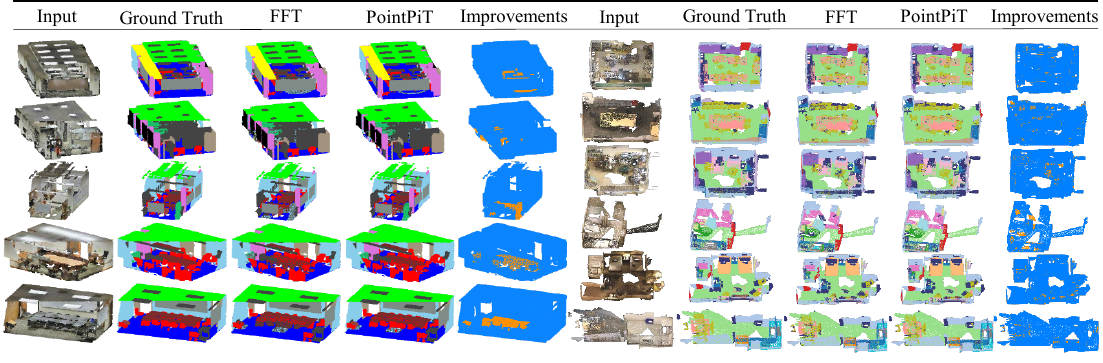} % Reduce the figure size so that it is slightly narrower than the column. Don't use precise values for figure width. This setup will avoid overfull boxes.
  \vspace{-20pt}
  \caption{Qualitative comparison of scene-level semantic segmentation. The left and right parts present results on S3DIS Area 5~\cite{Armeni_2016_CVPR} and ScanNet~\cite{dai2017scannet}, respectively.  Compared with FFT, PointPiT generates more accurate and
    consistent predictions in complex scene regions.}
  \label{fig3}
\end{figure*}

\begin{table*}[t]
  \small

  \begin{minipage}[t]{0.28\textwidth}
    \vspace{0pt}
    \centering
    \captionof{table}{Ablation study of different components of PointPiT.}
    \label{tab:abl_ssa_gso}
    \vspace{-6pt}
    \setlength{\tabcolsep}{3pt}
    \renewcommand{\arraystretch}{0.8}

    \begin{tabular}{c|c|c|cc}
      \toprule
      \multirow{2}{*}{SSA}    &
      \multirow{2}{*}{GSO}    &
      \multirow{2}{*}{Param.} &
      \multicolumn{2}{c}{ScanNet Val~\cite{dai2017scannet}}
      \\
      \cmidrule(lr){4-5}
                              &               &      & mAcc & mIoU
      \\
      \midrule
                              &               & 0.02 & 83.1 & 72.2 \\
      \checkmark              &               & 1.08 & 85.8 & 77.5 \\
                              & \checkmark    & 0.02 & 84.6 & 72.9 \\
      \rowcolor{tablegreen}
      \checkmark              & \checkmark    & 1.08 &
      \textbf{86.7}           & \textbf{78.9}
      \\
      \bottomrule
    \end{tabular}

  \end{minipage}
  \hfill
  \begin{minipage}[t]{0.4\textwidth}
    \vspace{0pt}
    \centering
    \captionof{table}{Sensitivity to different serialization strategies.
      We report mIoU (\%) on ScanNet Val~\cite{dai2017scannet}.}
    \label{tab:abl_serialization}
    \vspace{-6pt}
    \setlength{\tabcolsep}{3.5pt}
    \renewcommand{\arraystretch}{0.78}

    \begin{tabular}{l|cccc}
      \toprule

      Method
       & Z-order
       & Hilbert
       & Trans-Z
       & Trans-H
      \\

      \midrule
      Sonata (w/o dec.)
       & 78.9          & 77.6 & 78.1 & 78.2 \\
      \midrule
      Sonata (lin.)
       & 72.2          & 68.4 & 70.1 & 68.9
      \\
      +LoRA
       & 76.3          & 71.3 & 73.5 & 72.0
      \\
      +PointTPA
       & 78.4          & 73.9 & 76.8 & 74.6
      \\
      \rowcolor{tableblue}
      +PointPiT
       & \textbf{78.9}
       & \textbf{78.0}
       & \textbf{78.6}
       & \textbf{78.2}
      \\
      \bottomrule
    \end{tabular}

  \end{minipage}
  \hfill
  \begin{minipage}[t]{0.261\textwidth}
    \vspace{0pt}

    \captionof{table}{Overall analysis of key hyperparameters in PointPiT.}
    \vspace{-6pt}
    \label{tab:abl_hyperparameters}

    \scriptsize
    \setlength{\tabcolsep}{2pt}
    \renewcommand{\arraystretch}{0.914}

    \resizebox{\linewidth}{!}{%
      \begin{tabular}{cc|cc|cc}
        \toprule
        $r$       & mIoU          &
        $q$       & mIoU          &
        $\lambda$ & mIoU                                    \\
        \midrule
        16        & 76.5          & 8   & 77.6 & 0.0 & 78.1 \\
        32        & 78.0          & 16  & 78.4 & 0.1 & 78.6 \\
        \rowcolor{tablered}
        64        & \textbf{78.9} &
        32        & \textbf{78.9} &
        0.5       & \textbf{78.9}                           \\
        128       & 78.7          & 64  & 78.6 & 1.0 & 78.3 \\
        256       & 78.5          & 128 & 78.2 & 2.0 & 77.6 \\
        \bottomrule
      \end{tabular}%
    }

  \end{minipage}
  \vspace{-6pt}
\end{table*}

\section{EXPERIMENTS}
\subsection{Experimental Setup}
\vspace{-1pt}
\textbf{Datasets.}
To evaluate scene-level semantic segmentation, we benchmark PointPiT on
three large-scale benchmarks, including ScanNet~\cite{dai2017scannet},
ScanNet++~\cite{yeshwanthliu2023scannetpp}, and S3DIS~\cite{Armeni_2016_CVPR}.
These datasets contain diverse indoor scenes with different scales and
semantic categories, providing comprehensive evaluations for scene-level
adaptation.

\noindent
\textbf{Implementation Details.}
We adopt PTv3~\cite{wu2024pointtransformerv3}
initialized with pre-trained Sonata~\cite{wu2025sonata} weights as the
backbone. Following the PEFT setting~\cite{zha2023idpt}, the backbone
parameters are frozen during training, and only the proposed SSA modules and
task heads are optimized. we set the prefix~\cite{li2021prefix} token length to 4, the LoRA~\cite{hu2022lora} rank to 32, and the adapter's~\cite{houlsby2019parameter} intermediate dimension to 64. The hyperparameters are shared across all datasets throughout the validation process.
All methods follow the official downstream training and evaluation protocol. For fair comparison, all PEFT methods and the corresponding full fine-tuning
baseline use the same Sonata backbone without the decoder. All experiments are conducted on 8 NVIDIA A800 GPUs.

\noindent
\textbf{Evaluation Metrics.}
We report mean Intersection over Union (mIoU, \%), mean Accuracy (mAcc, \%),
and overall Accuracy (allAcc, \%) for semantic segmentation following prior
works~\cite{zhao2021pointtransformer,pang2022pointmae,qi2017pointnet,guo2026mantis}. We additionally report the number of trainable parameters to evaluate
the parameter efficiency of different adaptation methods.

\vspace{-3pt}
\subsection{Main Results}

\textbf{Quantitative Results.} As shown in Table~\ref{tab:scannet}, compared to full fine-tuning, PointPiT achieves competitive
performance and even exceeds it on ScanNet Val~\cite{dai2017scannet}
with less than 1\% of the backbone's parameters.
Across semantic categories, PointPiT attains the best or second-best results in most cases.
Furthermore, Table~\ref{tab:indoor_seg} demonstrates that
PointPiT consistently outperforms all representative
PEFT methods across diverse scene-level benchmarks.
These benchmarks exhibit substantial differences in geometric structure and semantic granularity, demonstrating the effectiveness of PointPiT across diverse scene-level datasets.

\noindent
\textbf{Qualitative Results.} As illustrated in Fig.~\ref{fig3}, PointPiT produces more accurate predictions than FFT in complex scene regions, with fewer semantic misclassifications and more coherent object structures. These results demonstrate its ability to preserve fine-grained semantics under parameter-efficient adaptation.

\vspace{-3pt}
\subsection{Ablation Study}

\textbf{Effect of SSA and GSO.}
Table~\ref{tab:abl_ssa_gso} demonstrates the complementary contributions of
SSA and GSO. Without SSA, GSO operates only on the task head
and yields a modest improvement. When combined with SSA, GSO further
improves performance, indicating the benefit of partition-aware
optimization over the adapted representations.

\noindent
\textbf{Sensitivity to Serialization Configuration.} Different serialization configurations result in distinct partition
structures, which may introduce unstable adaptation patterns for PEFT
methods. To isolate partition sensitivity, we evaluate the same checkpoint
on identical scenes under different test-time serializations. By capturing
more stable structural patterns and avoiding over-reliance on a specific
partition layout, PointPiT maintains consistent performance across
different configurations, as shown in Table~\ref{tab:abl_serialization}.
This suggests that the proposed adaptation strategy improves the
generalization ability of scene-level models under varying serialization
schemes.

\noindent
\textbf{Analysis of Key Hyperparameters.} Table~\ref{tab:abl_hyperparameters} shows that \(r=64\), \(q=32\), and \(\lambda=0.5\) achieve the best performance. Smaller \(r\) or \(q\) limits adaptation capacity or informative updates, while excessive values introduce diminishing returns. Similarly, overly strong partition regularization may suppress useful gradients. These results indicate that PointPiT achieves effective adaptation with moderate model capacity and selective gradient updates, without relying on excessive parameter expansion or partition regularization.

\section{Conclusion}
In this paper, we propose PointPiT,
a partition-invariant tuning framework for
large-scale scene-level point clouds. We
identify the mismatch between serialization-induced
partition variations and parameter-efficient adaptation,
and address it through scene-aware structural adaptation
and gradient subspace optimization. Extensive experiments
demonstrate that PointPiT achieves consistent state-of-the-art performance
among PEFT methods and competitive or even superior performance
to full fine-tuning with less
than 1\% of the backbone's parameters. As an early
exploration of scene-level PEFT, we expect PointPiT
to serve as a strong baseline and provide insights for future research on efficient 3D scene adaptation.

\bibliographystyle{IEEEbib}
\bibliography{strings,refs}

\end{document}